\documentclass[conference]{IEEEtran}
\usepackage{blindtext}
\usepackage{graphicx}
\usepackage{float}
\usepackage{array}
\usepackage{amsmath}
\usepackage[nottoc]{tocbibind}
\usepackage{graphicx}
\usepackage[utf8]{inputenc}

\usepackage{cite}
 
\usepackage[draft]{hyperref}

\begin{document}
\title{Cryptocurrency Portfolio Management with Deep Reinforcement Learning}

\author{\IEEEauthorblockN{Zhengyao Jiang}
\IEEEauthorblockA{Xi'an Jiaotong-Liverpool University \\
Email: zhengyao.jiang15@student.xjtlu.edu.cn}
\and
\IEEEauthorblockN{Jinjun Liang}
\IEEEauthorblockA{
Xi'an Jiaotong-Liverpool University \\
Department of Mathematical Sciences\\
Email: jinjun.liang@xjtlu.edu.cn}}


\maketitle

\begin{abstract}

	Portfolio management is the decision-making process
of allocating an amount of fund into different financial investment
products. Cryptocurrencies are electronic and decentralized
alternatives to government-issued money, with Bitcoin as the
best-known example of a cryptocurrency. This paper presents
a model-less convolutional neural network with historic prices
of a set of financial assets as its input, outputting portfolio
weights of the set. The network is trained with 0.7 years’ price
data from a cryptocurrency exchange. The training is done in a
reinforcement manner, maximizing the accumulative return,
which is regarded as the reward function of the network. Backtest
trading experiments with trading period of 30 minutes is
conducted in the same market, achieving 10-fold returns in
1.8 month’s periods. Some recently published portfolio selection
strategies are also used to perform the same back-tests, whose
results are compared with the neural network. The network is
not limited to cryptocurrency, but can be applied to any other
financial markets.
\end{abstract}


\section{Introduction}\label{introduction}



	Portfolio management is the decision making process of allocating an amount of fund into different financial investment products, aiming to maximize the return while restraining the risk \cite{haugen1986inv}\cite{markowitz1968portfolio}. Traditional portfolio management methods can be classified into four classes, "Follow-the-Winner", "Follow-the-Loser", "Pattern-Matching" and "Meta-Learning" \cite{li2014survey}.
The first two categories are based on prior-constructed financial models, while they may also be assisted by some machine learning techniques for parameter determinations \cite{Li2012}\cite{cover1996universal}. The performance of these methods is dependent on the validity of the models on different markets.
"Pattern-Matching" algorithms select part of history which is similar to current situation, and optimize the portfolio based on the selected history under some assumptions on the behavior of the market \cite{gyorfi2006nonparametric}.
The last class, "Meta-Learning" method, tries to combine multiple classes of methods to achieve better performance \cite{vovk1998universal}\cite{Das2011}.
In this work, we apply a full machine learning approach to the general portfolio management problem, without assuming any prior knowledge of the financial markets or making any models, and completely letting the algorithms observe and learn from the market history.


Many of the work applying deep machine-learning to financial market trading, tries to predict the price movements or trends \cite{Heaton2016}\cite{Niaki2013} using historic market data.
For example, with the input of a history price matrix, the network outputs a vector predicting the prices in the next period. This idea is straightforward because it is a case of supervised learning, and more percisely, regression problem.
Our trading robot does not, however, predict the price of any specific financial product, but directly outputs the market management actions, the portfolio vector. There are two reasons behind this design. The first reason is that trading actions, including what and how much to buy/sell in the market, based on predicted price movement will require human designed models to convert the latter to the former, and this is against our aim of a model-less trading algorithm. The second is high accuracy in predicting price movement is usually difficult to achieve, while the ultimate goal of portfolio management is to make higher profit instead of higher price-prediction accuracy.  
	Previous successful attempts of such model-less portfolio-selection machine learning scheme include some variants of \textit{Reinforcement Learning} (RL)\cite{cumming2013masterthesis}\cite{RRL}\cite{adaptive}. These algorithms output discrete singles, and make investments into single assets. Furthermore, they are limited to linear transformations, making them shallow learning.

	Existing deep reinforcement learning algorithms such as stochastic policy gradient based on probability models and deep Q-learning method \cite{Mnih2015}\cite{silver2016alphago}, making remarkable achievements in playing video and board games, are also limited to problems with discrete actions. In portfolio management problems, the actions are continuous. Although market actions can be discretized, discretization is considered a drawback. This is because discrete actions come with unknown risks. For instance, one extreme discrete action may be defined as investing all the capital into one asset, without spreading the risk to the rest of the market. In addition, discretization scales badly. Market factors, like number of total assets, vary. In order to take full advantage of adaptability of machine learning to different markets, trading algorithms have to be scalable.
	Another general deep reinforcement learning approach, called critic-actor Deterministic Policy Gradient, outputs continuous actions, training a Q function estimator as the reward function, and a second neural network as the action function\cite{Silver2014}\cite{Lillicrap2016}. Training two neural networks (the critic and the actor), however, is found out to be difficult, and sometimes even unstable.
In our approach, we employ a simple deterministic policy gradient using a direct reward function in the portfolio management problem, avoiding Q-function estimation.

Our trading algorithm is tested in a crypto-currency (virtual money, Bitcoin as the most famous example) exchange market, Polonix.com. A set of coins chosen by their previous trading-volume ranking are considered in the portfolio selection problem. Back-test trades are  performed in a period of 30 minutes. The performance of our back-test is compared that of three recent portfolio selection algorithms, summarised and implemented by Hoi \cite{li2014survey}, in the same cryptocurrency exchange.

	Cryptographic currencies, or simply cryptocurrencies, are electronic and decentralized alternatives to government-issued moneys \cite{nakamoto2008bitcoin}\cite{grinberg2012bitcoin}. While the best known example of a cryptocurrency is Bitcoin, there are more than 200 other tradable cryptocurrencies, called altcoins (meaning alternative to Bitcoin), competing each other and with Bitcoin \cite{bonneau2015sok}. The motive behind this competition is that there are a number of design flaws of Bitcion, and people are trying to invent new coins to overcome these defects hoping their inventions will eventually replace Bitcoin \cite{bentov2014proof}\cite{duffield2014darkcoin}. To November 2016, the total market capital of all cryptocurrencies is 13.8 billions in USD, 11.8 of which is of Bitcoin.\footnote{Crypto-currency market capitalizations, http://coinmarketcap.com/, accessed: 2016-11-25.} Therefore, regardless of its design faults, Bitcoin is still the dominant cryptocurrency in market. As a result, many altcoins can not be bought with fiat currencies, but only be traded against Bitcoin.

In this trading experiment, we do not consider the fundamental properties of cryptocurrencies, but only look at their technical aspects, namely price movement and volume. Two natures of cryptocurrenies, however, differentiate them from traditional financial assets, making their market the best test-ground for our novel machine-learning portfolio management experiments. These natures are decentralization and openness, and the former implies the latter. Without a central regulating party, anyone can participate in cryptocurency trading with low entrance requirements, and cryptocurrency exchanges flourish. One direct consequence is abundance of small-volumed currencies. Affecting the prices of these penny-markets will require smaller amount of investment, compared to traditional markets. This will eventually allow trading machines learning and taking the avantage of the impacts by their own market actions. Openness also means the markets are more accessible. Most cryptocurrency exchanges have application programming interface for obtaining market data and carrying out trading actions, and most exchanges are open 24/7 without restricting the number of trades. These non-stop markets are ideal for machines to learn in the real world in shorter time-frames.

This paper is organized as follow. Section \ref{problem} defines the portfolio management problem that we are trying to solve in this project.
Section \ref{data} introduces the data accessing and processing steps.
The core innovation of this paper, the deterministic policy gradient in portfolio management problem, will be described in section \ref{dpg}.
Section \ref{training} shows the training method of the network.
Section \ref{topology} demonstrates how hyperparameters are tuned and presents the final selected model parameters.
The final section \ref{performance} will evaluate the trading strategy by back-test.

\section{Problem Definition}\label{problem}
\subsection{Problem Setting}
Let $m$ number of assets selected to be traded, of which the prices for $n$ trading periods construct the \textit{global price matrix} $G$:

\begin{equation}
G=
\begin{pmatrix}
    x_{(1,1)}   & x_{(1,2)} & x_{(1,3)} & \dots & x_{(1,n)} \\
    x_{(2,1)}   & x_{(2,2)} & x_{(2,3)} & \dots & x_{(2,n)} \\
    \hdotsfor{5} \\
    x_{(m,1)}  & x_{(m,2)} & x_{(m,3)} & \dots & x_{(m,n)},
\end{pmatrix}
\label{eq:global}
\end{equation}
where $x_{(i,t)}$ is the price of $i$th asset at the beginning of the $t$th trading period. Each row of the matrix represents the price time-sequence of an asset. Specially, the first row is the riskless asset. For example, in our case, the riskless asset is Bitcoin whose price is always 1, and all the price is the exchange rate against Bitcoin. The $t$th column of the matrix is the price vector, denoted by $\vec v_t$, of $t$th trading period. 

By element-wise dividing $\vec v_{t+1}$ by $\vec v_{t}$, we get \textit{price change vector} of $t$th trading period $\vec y_t$:
\begin{equation}
\vec{y}_t := \vec{v}_{t+1} \oslash \vec{v}_t =
(\frac{ x_{(1,t+1)}}{x_{(1,t)}}, \frac{x_{(2,t+1)}}{x_{(2,t)}}, ..., \frac{x_{(m,t+1)}}{x_{(m,t)}} ).
\label{eq:y}
\end{equation}

Suppose an agent is investing on the market, and his investment on a trading period $t$ is specified by a \textit{portfolio vector} $\vec \omega_t =(\omega_{(t,1)},...,\omega_{(t,i)},...,\omega_{(t,m)})$, where $\omega_{(t,i)}$ represents the proportion of total capital invested in the $i$th capital, and thus $\sum_i \omega_{(t,i)} =1, \forall t$. In a portfolio management problem, the initial portfolio vector $\vec{\omega}_0$ is chosen to be the first basis vector in the Euclidean space, that is $\vec \omega_0=(1, 0, ...,0)$, indicating all the capital is in the riskless asset or in a fiat currency, before the first trading period. It is Bitcoin in our case.

If the transaction fee is ignored, the dot product of \textit{portfolio vector} $\vec{\omega}_t$ in the current period $t$, and the \textit{price change vector} $\vec{y}_t$ of the next, is the capital change rate $r_t$ (i.e. total capital in next period divided by that of this period) for the next trading period. 
\begin{equation}\label{eq:reward_t}
r_{t} =  \vec\omega_t \cdot \vec y_{t}.
\end{equation}

If the commission fee is $C$ per Bitcoin, the total transaction fee in $t$th trading period is then:
\begin{equation}
 \mu_t = C \sum_{i=1}^{m} {|\vec{\omega}_{(t-1,i)} - \vec{\omega}_{(t,i)}|.}
\end{equation}
In our scenario, $C = 0.0025$, the maximum commision rate at Poloniex.

After $n$ trading periods the \textit{portfolio value}, which is equal to initial portfolio value plus the total return,  $\alpha_n$ becomes:
\begin{equation}\label{eq:portfolio_value}
\begin{split}
     \alpha_n 
         &= \prod_{t=0}^{n} r_{t}(1 - \mu_t)  \\
         &= \prod_{t=0}^{n}{\vec\omega_{t} \cdot \vec y_{t} 
                (1 - C\sum_{i=1}^{m} {|\omega_{(t-1,i)} - \omega_{(t,i)}|})},
\end{split}
\end{equation}
where the unit of portfolio value is chosen such that $\alpha_0 = 1$.

At the beginning of each trading period $t$, the agent obtains $m$ sequences of history prices, and based on them, makes the investment decision, $\vec \omega_t$.
This process will repeat until the last trading period. The purpose of our algorithmic agent is to generate, in this process, a sequence of \textit{portfolio vector} $\{ \vec \omega_1, \vec \omega_2, ..., \vec \omega_n \}$ in order to maximize the accumulative capital. 

\subsection{Two Hypothesises}\label{section:hypothesises}
In this work, we only consider back-test trading, where the trading agent pretends to be back in time at a point in the market history, not knowing any "future" market information, and does paper trading from then onward. Therefore we impose the following two assumptions.
\begin{enumerate}
\item market liquidity: Each trade can be finished immediately at the last price when the orders are put.
\item capital impact: The capital invested by the algorithm is so insignificant that is has no influence on the market.
\end{enumerate}

\section{Data} \label{data}
The price data obtained from Poloniex is one year in time span and the trading period is half an hour. All data is constructed into a \textit{global price matrix} $G$ in (\ref{eq:global}).

The input of the CNN is an $m\times w$ price matrix in the $t$th trading period $X_t$, of which each row is the price sequence of a coin during last $w$ trading periods, a trading window.
In our experiment, $m = 12$ and $w =50$.

Another part of data required in the training process and performance evaluation is the \textit{price change vectors} in next trading periods $\vec y_{t}$ defined in (\ref{eq:y}).

\subsection{Coin Selection}
There are about 220 cryptocurrencies that can be invested at Poloniex. In this auto-trading strategy, 12 most-volumed assets are selected to be traded. 
The reason for selecting the top volumed coins is that bigger volume implies better market liquidity of the assets.
In turn it means the situation in reality will be closer to Hypothesis 1.
Higher volumes also mean that the investment can have less influence on the market, establishing an environment closer to the Hypothesis 2.
Considering relatively high trading frequency (30 minutes) compared to some daily trading algorithms designed for stock markets, volume size is particularly important in the current setting.

The market of cryptocurrency is not stable.
Some previously rarely-traded coins can have sudden boost or drop in volume in a short period of time.
Consequently, taking volume of longer time-frames, for example several days, will be a better choice of coin selection criterion than that of one trading period (30 minutes in this paper).

On the other hand, choosing the top volumes at current time may raise the \textit{ volume prediction problem}, which denotes that the selection process itself provide some future information on the test set to the agent. Although this problem seems minor, it will have influences on performance at the final trading experiments.
Therefore, the volume ranking used is based on the average of the 30 days before the beginning of each back-test time-slot.

\subsection{Data Preprocessing}
\subsubsection{Normalization}
The absolute price values of the assets in the problem are not important for the agent to make any trading decisions, but only changes in price matter. Therefore, input prices to the network are normalized, dividing the current price vector. For an input window of $w$ periods, we define a local normalized price matrix, or simply price matrix, feeding the neural network. This price matrix reads 
\begin{equation}\label{eq:pricematrix}
X_t =
\begin{pmatrix}
    \frac{x_{(1,t-w+1)}}{x_{(1,t)}}       & \frac{x_{(1,t-w+2)}}{x_{(1,t)}} & \dots & 1 
    \\
    \frac{x_{(2,t-w+1)}}{x_{(2,t)}}       & \frac{x_{(2,t-w+2)}}{x_{(2,t)}} & \dots & 1
    \\
    \dots& \dots& \dots& \dots \\
    \frac{x_{(m,t-w+1)}}{x_{(m,t)}}       & \frac{x_{(m,t-w+2)}}{x_{(m,t)}} & \dots & 1 
\end{pmatrix}.
\end{equation}

To train the network, we also need the price change vector of the period, $\vec{y}_t$, to define the reward function, which will be given in Section \ref{dpg}.

\subsubsection{Filling Empty History Data}
Some of the coins lack part of the history data, most of the lacking is because these coins just appeared recently. The data before the existence of a coin is marked as Not A Number (NAN). Normally, NANs only appeared in the training set, because the coin selection criterion is the volume-ranking of the last 30 days in the training set, meaning all assets must have existed before the back-test.

As the input of the CNN must be real numbers, these NANs should be replaced. The simplest apprach is to replace them with 1, indicating the price did not fluctuate before launching. However, during the training, it is meaningless to invest the nonexistent asset (but it's not cheating because there is the riskless asset, Bitcoin, in the assets set).
Moreover, this part of history may also be learnt by the CNN and such asset may be recognized as riskless. It is not expected that the algorithm to invest in coins which lacks a large part of history, because less training data means a higher probability of over-fitting.
Therefore, a fake decreasing price series is filled with decay rate 0.01 in the blank history for each coin if necessary, in order to prevent the agent from investing that asset. Note that the decay rate can not be set bigger than 0.05 or the training process will be easily trapped in local minima.

\subsection{Dividing Data into Three Sets}
The \textit{global price matrix} $G$ is divided into three parts, training, test, and cross-validation sets. The neural network will learn, in practice tuning the weights, in the training set. The test set can be used to evaluate the final performance of this algorithm comparing with other modern portfolio management algorithms. The cross-validation set is used to tune hyperparameters, such as the number of neurones in the hidden fully-connected layer of the network.
The ratio among these three sets is $0.7:0.15:0.15$.
\subsection{Perspective of Reinforcement Learning}
In the perspective of reinforcement learning, the total capital change after each trading period $r_t$, define in Equation (\ref{eq:reward_t}),  is the reward; the output \textit{portfolio vector} $\vec \omega_t$ is the action; and the history price matrix $X_t$ is used to represent the state of the market.
Therefore the whole portfolio management process of $n$ trading periods can be represented as a state-action-reward-state trajectory $\tau = ({X_1, \vec{\omega}_1, r_1, X_2, \vec{\omega}_2, r_2, ..., X_n, \vec{\omega}_n, r_n})$.
Note that under the hypothesises set in \ref{section:hypothesises}, the action $\vec{\omega}_t$ will not influence the state information in next period $X_{t+1}$.
As the experiment method being \textit{back-test}, which uses history data to mimic a real trading, can not provide such influence.

\section{Derterminstic Policy Gradient}\label{dpg}
\subsection{Portfolio Weight as Output}
Traditional ways of using CNN in financial is to predict the change in price, so the output is predicted price vector, common policy gradient networks output the probability of each action, limiting the action to discrete cases.
Different from these two approaches, our network directly outputs the portfolio weight vector, whose element is the ratio of total capital. For example, if the first element of the vector is 0.2, the algorithm will keep $20\%$ of the total capital in the first asset.
In this article, $\vec{\omega}$ is used to denote the portfolio weight vector.

\subsection{Reward Function}
The goal of the algorithm is to maximize the \textit{portfolio value} $\alpha$. Therefore, the reward function, or the objective function in supervised learning, is:

\begin{equation}
R_0= \sqrt[n]{\prod_{t=1}^{n-1}{\vec\omega_{t} \cdot \vec y_{t} (1 - C\sum_{i=1}^{m} {|\vec\omega_{(t,i)} - \vec\omega_{(t+1,i)}|})}} 
\end{equation}

As the input matrix does not include \textit{portfolio vector} $\vec \omega$ of the last period, adding the transaction cost term into the reward function will not be helpful but will slow down the training. Thus, this term is ignored. The reward in each period is taken logarithm for the sake of computational efficiency. The final reward function, the average logarithmic return, then looks:

\begin{equation}
	R=\frac{1}{n}\sum_{t=1}^{n+1}\ln{\vec{\omega_{t}} \cdot  \vec{y_{t}} }
\end{equation}

Note that, each portfolio vector $\vec{\omega}_t$ satisfies $\sum_i \omega_{t,i} = 1$.
To achieve this, \textit{softmax} is used as the activation function in the output layer.

\subsection{Advantages and Limitations}

Differing from the normal reinforcement learning algorithm, in which the action (output of the network) does not have explicit mapping relationship with the reward, our algorithm directly optimize the value of reward function without probability technique, preventing the high-variance problem and giving us the freedom to build the consistent action model.
This can also provide more extensibility and scalability, for example adding risk terms and volumes, compared to prediction based method.

Under our hypothesis that the investment of the agent will not affect the price of assets, the environment state will not be influenced by the actions of our agent.  More precisely, when taking the transaction fee into account, only the final return will be affected by our own trading volume.
Hence, the input of the network is not dependent on the last output, and the training method is not limited with the stochastic learning.
Furthermore, the form of the training can be more similar to the supervised learning and many tricks in supervised learning can be transplanted here.

\section{Network Training}\label{training}
The training process is to tune the weights of the Neural Network using gradient based methods to maximize the reward function on the training set.

The fact that the input of the agent is indepent of the last output, allows us to use the mini-batch training to speed up computation\cite{Li2014}.
The order of mini batches is shuffled in each epoch.

The initial values of the weights of the network, distributed normally with standard deviation of 0.1 and expectation of 0,  play an important role during the training.
The final performance on the cross-validation set varies a lot in different trials of training under same hyperparameters, indicating that the training is easily ending up with local minima.

Adam Optimization are used in training. The learning rate is $10^{-5}$ \cite{Kingma2014}.
Total Steps of the training is 900000.
Dropout with keep probability of 0.3 and L2 regularization with coefficient $10^{-8}$ are employed in order to prevent over-fitting \cite{Srivastava2014}.
Detailed hyper-parameters for training and data accessing are listed in Table \ref{tab:hyperparameters}.

\section{Network Topology}\label{topology}
\subsection{Model Selection}
Model selection is carried out on the global price matrix G.
Because of the local minimum problem, one set of initial weight values can not garentee to have the optimal result.
Instead, the network is trained with 5 to 8 sets of initial values. 
Hyper-parameters with the highest reward $R$ on the cross-validation are chosen.
In Table \ref{tab:cnn} and \ref{tab:fully-connected}, there are result data of training of two randomly selected CNNs with different topology, the standard deviation of performance on both cross-validation set and test set is high.

\begin{table}
\begin{center}
\begin{tabular}{|m{5em}|m{3.5em}|m{3.5em}|m{3.5em}|m{4em}|} 
 \hline
       & maximum & minimum & mean & standard deviation\\ 
 \hline
 CNN-1 test & 16.19 & 0.81 &  6.18 & 4.29 \\
 \hline
 CNN-1 CV & 18.07 & 1.77 & 4.9 & 5.02 \\
 \hline
 CNN-2 test & 16.21 & 3.08 & 5.98 & 4.01\\ 
 \hline
 CNN-2 CV & 5.58 & 2.41 & 3.32 & 0.99 \\
 \hline
\end{tabular}
\end{center}
\caption{Training Sample of CNN}
\label{tab:cnn}
\small{
	Portfolio value $\alpha$ on the test or crossvalidation set in 8 training trials with different initial values for two different CNNs. Standard deviation is high and there is a large gap between maximum and minimum value.
	Final training result is sensitive to the initial value of the weights.
}
\end{table}
\begin{table}
\begin{center}
\begin{tabular}{|m{5em}|m{3.5em}|m{3.5em}|m{3.5em}|m{4em}|} 
 \hline
       & maximum & minimum & mean & standard deviation\\ 
 \hline
 DNN-1 test & 8.85 & 3.59 & 5.72 & 1.70 \\
 \hline
 DNN-1 CV & 6.34 & 3.98 & 5.09 & 0.79 \\
 \hline
 DNN-2 test & 8.64 & 0.62 & 5.27 & 2.47\\ 
 \hline
 DNN-2 CV & 6.41 & 1.63 & 4.76 & 1.64 \\
 \hline
\end{tabular}
\end{center}
\caption{Training Sample of Fully-connected Network}
\label{tab:fully-connected}
\small{
	Portfolio value $\alpha$ in 8 training trials of two different Fully-Connected Neural Networks.
	Standard deviation is smaller than that of CNNs while the best scores are lower than CNN's.
}
\end{table}

\subsection{CNN Topology}
\begin{figure}[h]
    \begin{center}
    \includegraphics[width=0.9\linewidth]{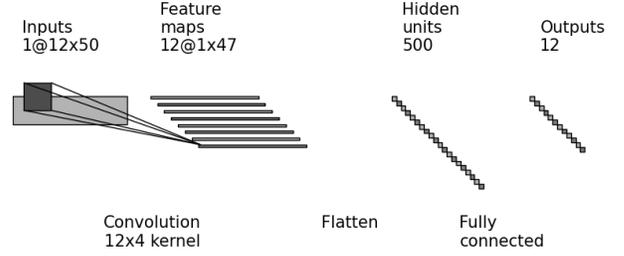}
    \caption{Diagram of our CNN topology}
    \label{fig:cnn_topology}
    \end{center}
    \small{
	    In the convolution layer, f@r $\times$ c means there is f features, each of them has r rows and c columns.
	    Therefore, the inputs of the network are 12$\times$50 matrices, which represent the prices of 12 coins in last 50 trading periods.
	    The output layer has 12 neurons, outputing the \textit{portfolio vector} $\omega$.
	    The activation function of all the hidden layers is ReLU and that of the output layer is softmax.
	}
\end{figure}

As a result of model selection, the best performed CNN has 2 hidden layer: a convolution layer and an fully-connected Layer as demostrated in the Fig. \ref{fig:cnn_topology}.
The height of filter is the number of coins or just one.
The reason is that in 2D-input situation like image recognition, there is local correlation between both adjacent rows and columns.
However, the order of the rows in input matrix $X$ of our CNN agent is arbitrary; thus, different rows are treated as different channels of the input, just like the color channels 'RGB' in the computer vision tasks. 
Our final choice of filter size is $12 \times 4$, therefore this convolution process could also been seen as 1 dimensional convolution.
The number of the filters is 12.
Pooling is not applied after the convolution layer. Pooling trades the translation invariance and reduction of parameters with loss of location information. 
Location in the price matrix is specifying time entry and coin type in the trading window, and hence is important.
Following the convolution layer there is a fully connected layer of 500 neurons, and a softmax output layer with 12 neurons.
The activation function of all the hidden layers is rectified linear unit (ReLU). 

Deeper structures with more convolution layers or more fully-connected layers have been tried, but none of them outperforms the one in Figure \ref{fig:cnn_topology}.
One of the reason might be the price movement only provides noisy information to the market state,
therefore complex topology will lead to over-fitting.
Failure of deeper structure may also caused by the small scale of training data, with only 12,000 data points. 
The training set is not extended to the older time, because cryptocurrency market are new, and most of the selected coins did not have such a long history.
Moreover, the training data that far away may have much less correlation with the test and cross-validation sets.
Further evidence can be seen in the Section \ref{performance}. 

\subsection{Fully Connected Network}
Results in Table \ref{tab:fully-connected} is from the best performing Fully Connected Neural Network. The performance is more stable than CNN while the best performance is worse.

\section{Performance Evaluation}\label{performance}
\subsection{Results} 
The \textit{back-test} experiments are executed with \text{global price matrices} $G$'s of time-spans 2015/06/27-2016/06/27, 2015/07/27-2016/07/27 and 2015/08/27-2016/08/27.
Moving the time-span of the global price matrix will not only move the time of the test set, but also the training and the cross-validation sets.
Hence, in the three experiments, the networks are trained, respectively, on 2015/06/27-2016/03/14, 2015/07/27-2016/04/14 and 2015/08/27-2016/95/27.
Then the best network of each experiment would be selected according to the performance on cross-validation set.
Final backtests are conducted on 2016/03/14-2016/05/03, 2016/04/14-2016/06/03, 2016/05/14-2016/07/03.

The performances of three benchmarks and three recent portfolio management algorithms, summaried by Li et.\ al.\cite{Li2012}, are compared with our CNN agent.
The first benchmark \textit{Uniform Buy and Hold} is a strategy investing wealth uniformly on each assets and holds the portfolio until the end.
The \textit{Best Stock} is the price movement of the asset that has the greatest increase in value during the abserved period.
The \textit{Uniform Constant Rebalanced Portfolio} is a baseline strategy which will rebalance the portfolio uniformly every trading period \cite{constantRebalance}.
The three portfolio algorithms are \textit{Universal Portfolio} \cite{cover1996universal}, \textit{Online Newton Step} \cite{Das2011} and \textit{Passive Aggressive Mean Reversion} \cite{Li2012}.
The commission rate in the back-test is 0.0025.

Besides the final portfolio value and standard deviation of returns for each period, two financial measures, Sharpe ratio and maximum drawdown, are used to evaluate the risk of strategies.
Sharpe ratio \cite{sharpe1964} is a measure of risk-adjusted return, defined as $S = \frac{\overline{r}_p - r_f}{\sigma_p}$, where $\overline{r}_p$ is the expected portfolio return, $r_f$ is the risk free rate of return (0 in this case), and $\sigma_p$ is the standard deviation of the portfolio value.
The second measure, maximum drawdown \cite{blanchard1979MaxDD}\cite{magdon2003maximum}, is the maximum loss from a peak to a trough of a portfolio, before a new peak is attained.

\begin{figure*}[h]
\begin{center}
\includegraphics[width=0.9\linewidth]{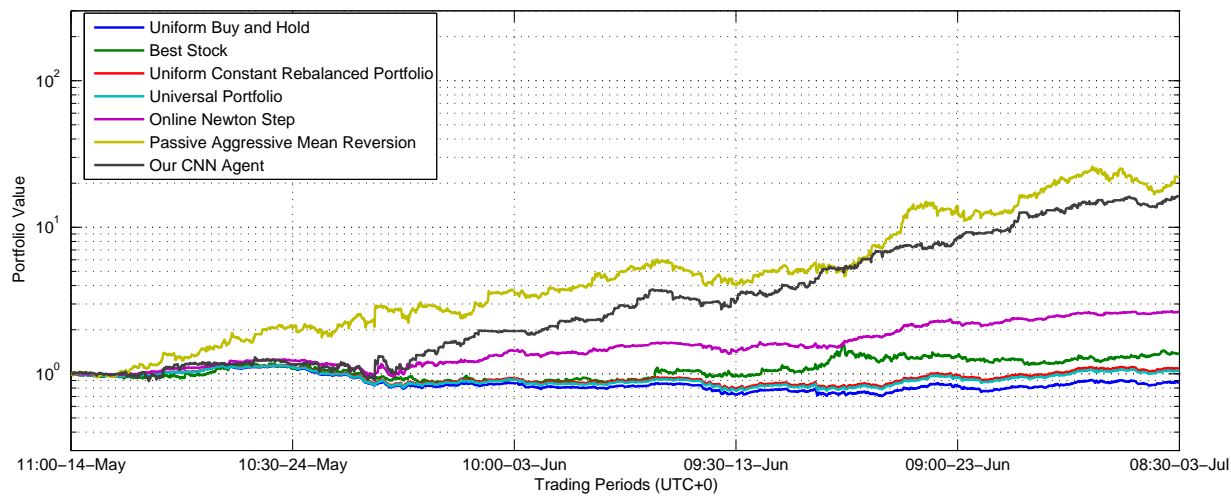}
\caption{Back Test Results}
\label{fig:result}
\end{center}
\small{
	Simulated trades are conducted from 2016/05/14 to 2016/07/03.
	The \textit{global price matrix} $G$ define in Equation \ref{eq:global} ranges from 2015/08/27 to 2016/08/27. 
	3 benchmarks and 3 other portfolio management methods are compared. Note that the vertical aixs is logarithmic.
	Our CNN agent over-performs, in term of final return, all the benchmarks and two of the other algorithms, only being second to the Passive Agressive Mean Reversion algorithm (PAMR). However, the overall risk of our CNN agent is smaller than that of PAMR, achieving a higher Sharpe ratio.
}
\end{figure*}

\begin{table*}
	\begin{center}
		\begin{tabular}{||m{17em}|m{6em}|m{6em}|m{6em}|m{6em}||} 
			\hline
			Algorithm Name& Final  Portfolio Value & Sharpe Ratio & Maximum Drawdown & Standard Deviation\\ 
			\hline
			Uniform Buy and Hold & 0.876067 & -1.541388 & 0.382002 & 0.003196 \\
			\hline
			Best Single Asset & 1.377662 & 1.125772 & 0.288306 & 0.001517 \\
			\hline
			Uniform Constant Rebalanced Portfolio & 1.090687 & -0.845601 & 0.316140 & 0.006255\\ 
			\hline
			Universal Portfolio & 1.048449 & -1.011029 & 0.330996 & 0.003528 \\
			\hline
			Online Newton Step & 2.648263 & \textbf{1.045801} & \textbf{0.278755} & 0.016226 \\
			\hline
			Passive Aggressive Mean Reversion & \textbf{21.872857} & 0.006260 & 0.353000 & 0.002559 \\
			\hline
			Our CNN Agent & \textbf{16.305332} & 0.036886 & 0.296082 & 0.002778 \\
			\hline
		\end{tabular}
	\end{center}
	\caption{Performance measures of algorithmic traders and benchmarks}
	\small{
	   These performance measures are for the back-test during 2016/05/14 - 2016/07/03.
	   The "Follow the winner" algorithm \textit{Universal Portfolio} gains the least return of the four strategies.
	   The "Meta-Learning" method \textit{Online Newton Step}, "Follow-the-Loser" method \textit{Passive Aggressive Mean Reversion} (PAMR)
	   and our CNN Agent performs well, gaining final portfolio value larger than the best single asset of all.
	   In terms of risk, the \textit{Online Newton Step} has the largest Sharp ratio and smallest max-drawdown, indicating this
	   algorithm is most stable.
	   Final portfolio values of the PAMR and our CNN agent are notably higher (20 and 15). While the former has a higher return, it has a greater risk.
    }
	\label{tab:result}
\end{table*}

\begin{figure*}[h]
\begin{center}
\appendix
\bigskip
\par
a. test set 2016/03/14-2016/05/03
\includegraphics[width=0.9\linewidth]{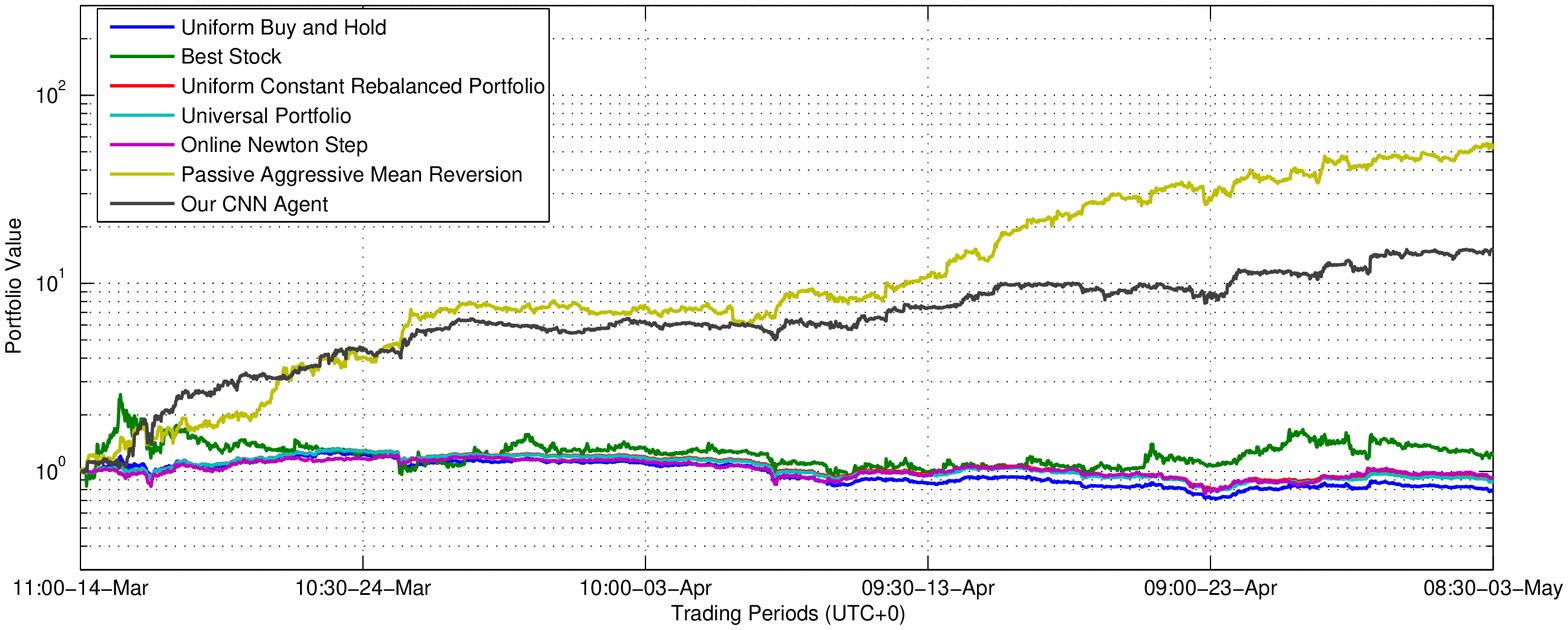}
\par
b. test set 2016/04/14-2016/06/03
\includegraphics[width=0.9\linewidth]{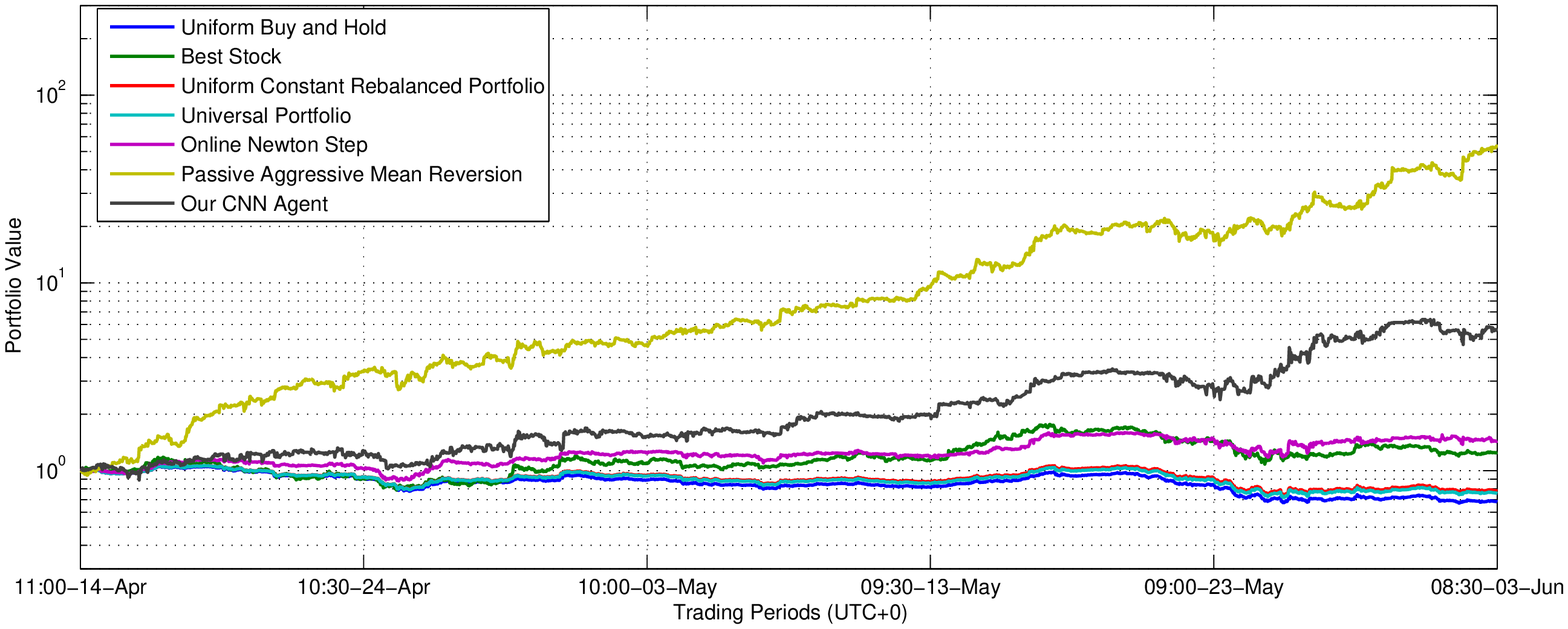}
\par
\end{center}
\caption{Back Test Results On Another two Test Set}
\label{fig:result2}
\small{
	Simulated trades are conducted for two other time-slots 2016/03/14-2016/05/03 and 2016/04/14-2016/06/03.
}
\end{figure*}

The result in the Fig. \ref{fig:result}, \ref{fig:result2} and Table \ref{tab:result} shows that the performance of our CNN agent outperforms most of the benchmarks and other compared algorithms, only losing to Passive Aggressive Mean Reversion, in term of accumulative return. However, our CNN trader achieves a signaficantly lower risk, resulting a higher Sharpe ratio than PAMR.

\subsection{The Expiration Problem}
Learning hidden market patterns from experience in the training set, the agent makes future decisions.
This is based on an assumption that some of the market patterns still work out of the training set. However, if the time interval between the training set and the beck-test is too long, some these learnt patterns may not longer be valid.

As shown in the results, the CNN performance of a back-test closer to the training set is better than the further one. This suggests that the validity of CNN trader algorithm is not without an expiratory duration.
If the agent is going to start a real online trading, it is wise to put the training set at the closer time to the current time, or even do online training while trading.
Back-tests are put closer to the training set, making its performance comparison to other algorithms more convincing.

\subsection{Dilemma Between Performance Evaluation and Hyperparameters tuning}
As mentioned in the previous section, the performance of our network is strongly depended on the time location of the training data set. 
Due to this constrain, there is a dilemma between the choices of performance evaluation and hyperparameters tuning.
\begin{figure}[h]
    \begin{center}
    \includegraphics[width=\linewidth]{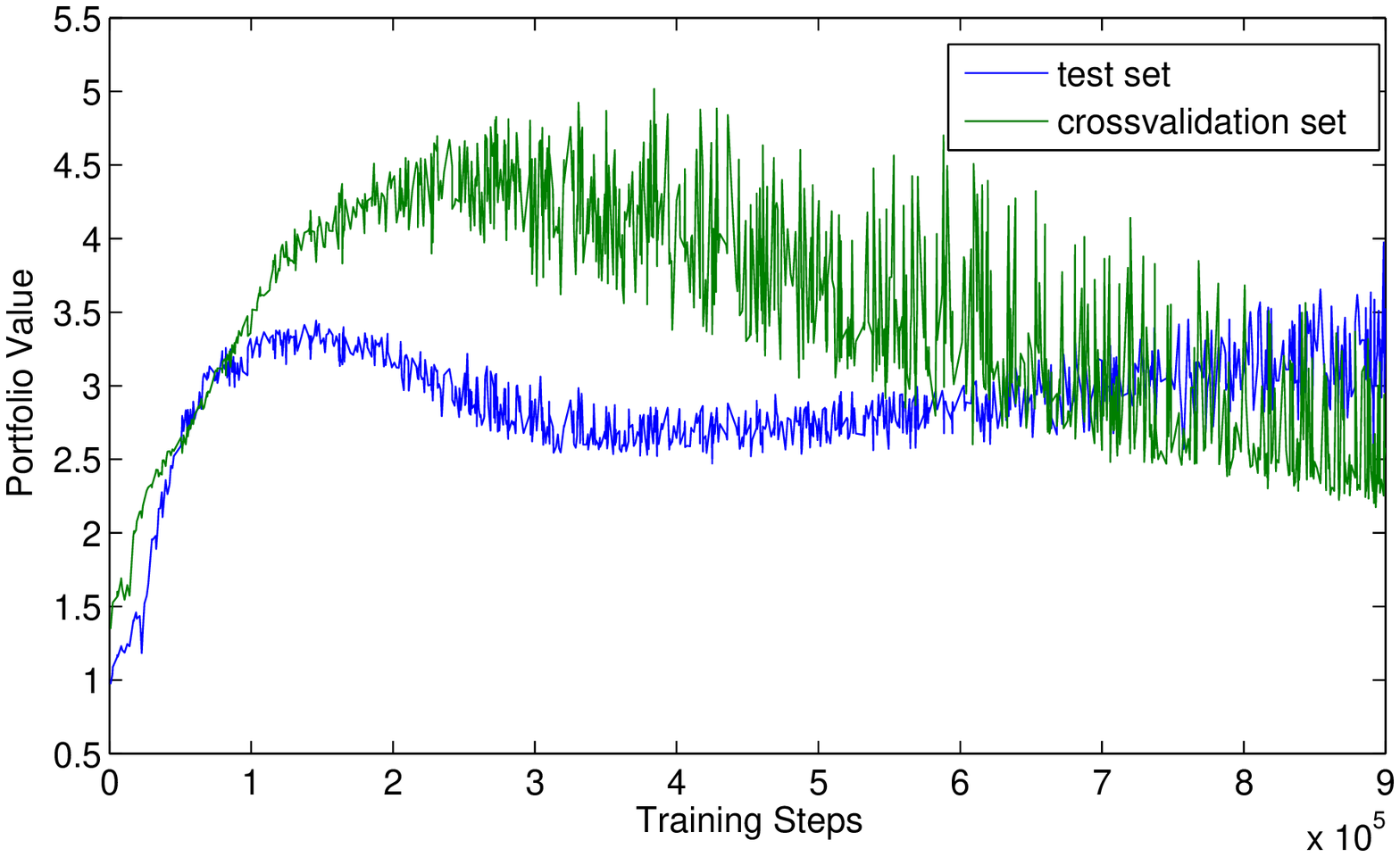}
    \caption{Training Process}
    \label{fig:train}
    \end{center}
	\small{
    	This picture shows the performance variation on the cross-validation set and test set during training.
		The horizontal axis shows the training epochs and vertial axis is the \textit{portfolio value}, defined in Equation (\ref{eq:portfolio_value}). 
    	When over-fitting happens in the cross validation set, the performance on test set is still going up.
	}
\end{figure}
As shown in the Fig. \ref{fig:train}, over-fitting happens at different epochs on the cross-validation  and test sets.

Suppose there are two types of hidden patterns in the market. One lasts longer than the other. 
When the target set is closer to the training set, both long-term patterns and short-term patterns work.
Whereas if the target set is far away to the training set, only the long-term patterns will work, and the short-term one learnt by the agent become an over-fitting factor.

Selecting hyper-parameters on the further set means the hyperparameters tend to suppress over-fitting, but may also ignore some short-term patterns that are useful in the test set.

\section{Conclusion}\label{conclusion}
In this article, we proposed a deterministic deep reinforcement learning method addressing the portfolio management problem, which directly produces the \textit{portfolio vector} $\vec \omega$ with raw market data, historic prices, as the input.
Our approach does not rely on any financial theory, therefore it is highly extensible.
A back-test experiment is carried out on a cryptocurrency market. The performance of the CNN strategy is compared with 3 benchmarks and 3 other portfolio management algorithms, achieving positive results. However, our method has a less cumulated return than the PAMR method.

The major limitation of this work is the training and testing of the algorithm is based on the two assumptions because we cannot use the history data to completely simulate the real on-line trading.
Furthermore, the cross-validation set is put at the end part of the \textit{global price matrix} $G$, which actually in the future of the test set.
If this method need to be applied in real market, we must think another way to do the model selection.

Another point that could be improved in the future is the training set is small and market single is limited; therefore, it is difficult to build a deeper network structure.

\bibliographystyle{IEEEtran}
\bibliography{reference}

\begin{table*}[b]
\begin{tabular}{||c|c|m{40em}||} 
 \hline
       hyperparameters & value & description\\ 
 \hline
 batch size & 50 & Size of mini-batch during training.  \\
 \hline
 window size & 50 & Number of the columns (number of the trading periods) of the input price matrices.  \\
 \hline
 number of coins & 12 & Total number of the assets (including Bitcoin) selected to be traded.  \\ 
 \hline
	trading period (second)& 1800 & Time interval of two trades, of which the unit is second.\\
 \hline
 fake decay rate& 0.01 & Faked price decay if the price is missing.\\
 \hline
 keep probability & 0.3 & Probability of a neuron is kept during dropout.\\
 \hline
 total steps & 900000 & Total steps of training.\\
 \hline
 regularization rate & $10^{-8}$ & Coefficient of the L2 regularization applied on the network while training.\\
 \hline
 learning rate & $10^{-5}$ & Parameter $\alpha$ (i.e. the step size) of the Adam optimization.\\
 \hline
	global time span (year) & 1 & Years of the time span of the \textit{global price matrix} $G$ in Equation (\ref{eq:global}), including the training set, cross-validation set and the test set.\\
 \hline
 training set portion & 0.7 & Time portion of the training set of the \textit{global price matrix} $G$.\\
 \hline
 cross-validation set portion & 0.15 & Time portion of the cross-validation set of the \textit{global price matrix} $G$.\\
 \hline
 test set portion & 0.15 & Time portion of the test set of the \textit{global price matrix} $G$.\\
 \hline 
 volume average days & 30 & Days of the total volume to be accumulated, which is the criterion to select assets to trade.\\
 \hline
	commision fee (per BTC)& 0.0025 &  Ratio of the capital that is consumed during a trading.\\
 \hline
 
\end{tabular}

\caption{Hyperparameters of the CNN agent.}\label{tab:hyperparameters}
\end{table*}

\end{document}